\documentclass[letterpaper, 10 pt, conference]{ieeeconf}  

\IEEEoverridecommandlockouts                              

\usepackage{graphicx} 
\usepackage{tensor}
\usepackage[font=footnotesize]{caption}
\usepackage{cite}
\usepackage[font=normal]{subcaption}
\usepackage{booktabs}
\usepackage{times} 
\usepackage{bm}
\usepackage{amsmath,amssymb,mathrsfs,mathdesign,upgreek,dsfont,gensymb}
\usepackage[ruled,vlined]{algorithm2e}
\usepackage{verbatim}
\usepackage{diagbox}
\usepackage{cleveref}
\usepackage{todonotes}
\usepackage{soul}
\usepackage[multiple]{footmisc}
\usepackage{multirow}
\usepackage{threeparttable}

\graphicspath{{./figures/}}

\DeclareCaptionSubType*[Alph]{table}
\DeclareCaptionLabelFormat{mystyle}{Table~\bothIfFirst{#1}{ ̃}#2}
\def\code#1{\texttt{#1}}
\DeclareMathOperator*{\argmin}{arg\!\min}

\newcommand*\diff{\mathop{}\!\mathrm{d}}

\title{\LARGE \bf
Bayes Net based highbrid Monte Carlo Optimization \\ for Redundant Manipulator
}

\author{Feng Yichang, Wang Jin$^{*}$, Zhang Haiyun, Lu Guodong 
\thanks{$^{*}$ Wang Jin, corresponding author,  1) State Key Laboratory of Fluid Power and Mechatronic Systems, School of Mechanical Engineering,
Zhejiang University, Hangzhou 310027, China;  2) Engineering Research Center for Design Engineering and Digital Twin of Zhejiang Province, School of Mechanical Engineering,
Zhejiang University, Hangzhou 310027, China. 
        {\tt\small dwjcom@zju.edu.cn}}%
}

\begin{document}

\maketitle
\thispagestyle{empty}
\pagestyle{empty}

\begin{abstract}

This paper proposes a Bayes Net based Monte Carlo optimization for motion planning (BN-MCO). Primarily, we adjust the potential fields determined by goal and start constraints to guide the sampled clusters towards the goal and start point progressively. Then we utilize the Gaussian mixed modal (GMM) to perform the Monte Carlo optimization, confronting these two non-convex potential fields. Moreover, Kullback–Leibler divergence measures the bias between the true distribution determined by the fields and the proposed GMM, whose parameters are learned incrementally according to the manifold information of the bias. In this way, the Bayesian network consisting of sequential updated GMMs expands until the constraints are satisfied and the shortest path method can find a feasible path. Finally, we tune the key parameters and benchmark BN-MCO against the other 5 planners on LBR-iiwa in a bookshelf. The result shows the highest success rate and moderate solving efficiency of BN-MCO. 

\end{abstract}

\section{INTRODUCTION}

The robot plays an important role in industrial automation and intelligence to improve the manufactory efficiency and to liberate more productive forces. Motion planning takes over their execution part to roll out safely and smoothly under some specific task requirements. So smoothness, collision free and goal reachability become the basic objectives of motion planning. 

To modify these objectives, some studies~\cite{Zucker2013CHOMP, Schulman2013SCO, Mukadam2018GPMP} use the numerical optimization to plan an optimal motion informed by the local manifold information nearby an initial trajectory. It means only when the initial is located in a convex hull nearby the optimum, does the local minimum, i.e. the failure case, be avoided. So other optimization based methods introduce the stochastic process~\cite{Kalakrishnan2011STOMP, Osa2020SMTO, Kuntz2020ISIMP} or machine learning~\cite{Ichnowski2020GOMP} during the optimization to place the trajectory in the convex hull. However, the frequent stochastic steps among variant hulls, originating from the high dimensional objectives with strong non-convexity, also worsen the optimization. Similarly, learning this kind of objectives also needs a large amount of modals or a complex network with deep layer structure, urging for a big dataset and long training time. 

Unlike the above stochastic steps, the sampling based methods employ random search and dynamic programming methods to find the feasible trajectories. Though some of them~\cite{Kavraki1998PRMs, LaValle1998RRTs, Kuffner2000RRT-connect, Karaman2011RRT*-PRM*} are proven probabilistic complete and perform efficiently facing the complex 2D and 3D maze scenarios, the number of sampled nodes and estimated edges of their searching graphs expands exponentially when the dimension of configuration space increases. So their exploitation ability as well as their success rate descends given a limited planning time in a complex scenario with massive obstacles surrounding the optimal trajectory. Some recent studies introduce the machine learning technology to accelerate collision check~\cite{Das2020iSVM} or to improve the searching efficiency through constructing the knowledgable manifold and its nearby region~\cite{Osa2022NN-TO} due to the planning experience. However, the offline unsupervised learning also consumes extra time. 

This paper treats the motion planning, a.k.a. goal approaching, as an optimization process like the potential field methods~\cite{Schulman2013SCO, Levenberg1944LM}. To avoid the local minima confronted by the numerical potential field, our method learns the multi-modal via the sequential Monte-Carlo sampling of waypoints in the configuration space. Unlike the above multi-modal method which samples the trajectories sequentially, our method samples serial bunches of waypoints towards the null space of the constraints in a potential field adjusted progressively. To save the computation resources for offline learning, we adopt the online learning to incrementally minimize the KL-divergence between the true distribution of the potential field and the mixture distribution defined by the multi-modals. The learned mixture distribution generates new waypoints used for the next learning until some of them satisfy the constraints. The construction of Bayes net for path searching synchronizes with the online learning, whose node is one single modal and the directed edge connects the last and learned modals. Since each modal consists of a bunch of related waypoints, our method can significantly reduce the amount of nodes and cut the redundant edges to improve the searching efficiency. 


\section{RELATED WORKS}

\subsection{Optimization based Motion Planning}
The main concern of numerical optimization is rapidly descending to an optimum.  CHOMP~\cite{Zucker2013CHOMP}, ITOMP~\cite{Park2012ITOMP}, and GPMP~\cite{Mukadam2018GPMP} adopt the gradient descent method with the fixed step size. CHOMP uses Hamiltonian Monte Carlo (HMC)~\cite{Shirley2011HMC} for success rate improvement. To lower the computational cost, GPMP and dGPMP~\cite{Bhardwaj2020dGPMP} adopt iSAM2~\cite{Kaess2012iSAM2} to do incremental planning, and each sub-planning converges with super-linear rate via Levenberg–Marquardt (LM)~\cite{Levenberg1944LM} algorithm. Meanwhile, ITOMP interleaves planning with task execution in a short-time period to adapt to the dynamic environment. Moreover, TrajOpt~\cite{Schulman2013SCO} uses the trust-region~\cite{Byrd2000TrustRegion} method to improve efficiency. It is also adopted by GOMP~\cite{Ichnowski2020GOMP} for grasp-optimized motion planning with multiple warm restarts learned from a deep neural network. Instead of deep learning, ISIMP~\cite{Kuntz2020ISIMP} interleaves sampling and interior-point optimization for planning. Nevertheless, the above methods may converge to local minima when the objective function is not strongly convex. 

\subsection{Sampling based Motion Planning}
Unlike the numerical method, the sampling method constructs a search graph to query feasible solutions or iteratively samples trajectories for motion planning. 

PRM~\cite{Kavraki1998PRMs} and its asymptotically-optimal variants like PRM*~\cite{Karaman2011RRT*-PRM*} and RGGs~\cite{Solovey2018RGGs} make a collision-free connection among the feasible vertexes to construct a roadmap. Then they construct an optimal trajectory via shortest path (SP) algorithms like Dijkstra~\cite{Dijkstra1959Dijkstra-Alg} and Chehov~\cite{Hofmann2015Chehov}, which store and query partial trajectories efficiently. Unlike PRM associated with SP, RRT~\cite{LaValle1998RRTs} and its asymptotically-optimal variants like RRT*~\cite{Karaman2011RRT*-PRM*} and CODEs~\cite{Rajendran2019CODES} find a feasible solution by growing rapidly-exploring random trees (RRTs).

STOMP~\cite{Kalakrishnan2011STOMP} resamples trajectory obeying Gaussian distribution renewed by the important samples. \cite{Petrovic2019HGP-STO} introduces the GPMP's function to improve the searching efficiency of STOMP. Moreover, SMTO~\cite{Osa2020SMTO} applies Monte-Carlo optimization for multi-solution and refines them numerically.

\section{PROBLEM FORMULATION}

\subsection{Potential field}\label{sec:costs}
This paper treats the trajectory/path searching as an approaching progress, BN-MCO adopts the artificial potential field (PF) to drive the waypoints towards the null space of constraints. Unlike RRT\cite{LaValle1998RRTs} growing the random searching tree from the initial waypoint towards the goal constraint, BN-MCO extends the Bayes net from the initial multi-modal informed by the same bunch of waypoints in bi-direction, i.e. forwards the goal and backwards the start, for a wider exploration. Therefore, we define the forward PF
\begin{equation} \label{eq:forwardPF}
	\mathcal{F}_{goal}(\bm{\theta}) =  f_{goal}(\bm{\theta}) + \lambda_c \cdot c(\bm{\theta}) +  b(\bm{\theta})
\end{equation}
and the backward PF
\begin{equation} \label{eq:backwardPF}
	\mathcal{F}_{start}(\bm{\theta}) =  f_{start}(\bm{\theta}) + \lambda_c \cdot c(\bm{\theta}) +  b(\bm{\theta}),
\end{equation}
where $f_{goal}$  and $ f_{start}$ penalize the goal and start constraints respectively to provide attraction forces, $c(\bm{\theta})$ with the magnifier $\lambda_c$ repulses the waypoints from the in-collision/unsafe area and $b(\bm{\theta})$ checks the joint limits. 

Considering the convergence of numerical optimization, some traditional PF methods add the kinetic energy part $\mathcal{K}(\bm{\theta},\bm{\theta}_{t-1}) = \|\bm{\theta} - \bm{\theta}_{t-1}\|^2_{\mathbf{A}}$ with a positive-definite matrix $\mathbf{A}$ to ensure the convexity of objective. \Cref{eq:forwardPF-convex} shows how to generate a series of waypoints $[\theta_{\frac{T}{2}+1}, \dots, \theta_{T}]$ incrementally from time $T/2$ to $T$.  
\begin{align} 
	\label{eq:forwardPF-convex}
	\bm{\theta}_{t} = \argmin_{\bm{\theta}} \mathcal{F}_{goal}(\bm{\theta}) + \mathcal{K}(\bm{\theta}, \bm{\theta}_{t-1}), 
	t \in (T/2,T], 
\end{align}
Since BN-MCO searches a feasible path in bi-direction from the initial modals, \cref{eq:backwardPF-convex} shows how to construct the backward trajectory $[\bm{\theta}_{\frac{T}{2}-1}, \dots, \bm{\theta}_0]$ from $T/2$ to $0$.
\begin{align}
	\label{eq:backwardPF-convex}
	\bm{\theta}_{t-1} = \argmin_{\bm{\theta}} \mathcal{F}_{start}(\bm{\theta}) + \mathcal{K}(\bm{\theta}, \bm{\theta}_{t}), 
	t \in (0,T/2]
\end{align}
\Cref{sec:GMM,sec:findPath} will detail how to learn the mixture distribution to build the forward and backward Bayes nets and how they compound for a bidirectional search of an optimal trajectory $\bm{\xi} = [\bm{\theta}_0, \dots, \bm{\theta}_{\frac{T}{2}},\dots,\bm{\theta}_{T}]$, correspondingly.

\subsection{Constraint construction}

\subsubsection{Goal constraint}
Since this paper predefines the goal constraint as $\bm{\Theta}_{goal} = \{\bm{\theta} | \bm{x}^{\min} \leq \bm{x}(\bm{\theta}) \leq \bm{x}^{\max}\}$ to restrict the goal state of end effector $\bm{x}(\theta)$ under the configuration $\theta$. To ensure the forward PF \eqref{eq:forwardPF} semi-positive, we adopt $\ell_2$ penalty  $(|a|^{+})^2 = \max(a,0)^2$ of the inequality constraint $a \leq 0$: 
\begin{equation} \label{eq:goalConstraint}
	f_{goal}(\bm{\theta})  =  \sum_{i = 1}^{6}\lambda_i [(|{x}^{\min}_i - {x}_i|^{+})^{2} + (|{x}_i-{x}^{\max}_i|^{+})^{2}],
\end{equation}
where the state of end-effector $\bm{x}(\bm{\theta}) = [x_1,\dots,x_6]^\text{T}$ is calculated via the forward kinematics of a robot with the joint state (i.e., configuration $\bm{\theta}$), $\{x_1, x_2, x_3\}$ and $\{x_4, x_5, x_6\}$ are the posture and position of end effector state respectively, and $\bm{\lambda} = [\lambda_1,\dots,\lambda_6]$ are the penalty factors. Considering the convexity of different constraints $i$, \Cref{sec:parameters} details the adjustment of their corresponding factor $\lambda_i$ for implementation. 

\subsubsection{Start constraint}
Since the start configuration $\theta_0$ is determined, we use a symmetric semi-positive definite matrix $\mathbf{M}$ to calculate the distance between a waypoint  $\bm{\theta}_0$ and $\bm{\theta}$:
\begin{equation}
	f_{start}(\bm{\theta}) = (\bm{\theta} - \bm{\theta}_0)^\text{T}\mathbf{M}(\bm{\theta} - \bm{\theta}_0), 
\end{equation}
so that the mixed models will be attracted backwards the start and \Cref{sec:parameters} will detail the selection of $\mathbf{M}$. 

\subsection{Collision check}\label{sec:obs}
Since the collision-free trajectory means a safe motion or successful planning, we first utilize $\bm{x}_{i,\tau} = \bm{x}(\mathcal{B}_i, \tau)$ to map from a state $\theta_\tau$ at time $\tau$ to the position state of a collision-check ball (CCB-$\mathcal{B}_i$) on the manipulator. Then we calculate the collision cost 
\begin{equation} \label{eq:collisionCost}
	c(\bm{x}_{i,\tau}) = \left \{
	\begin{array}{cl}
		0 & d(\bm{x}_{i,\tau}) > \epsilon \\
		\frac{1}{2\epsilon}[d(\bm{x}_{i,\tau}) - \epsilon]^2 & 0 \leq  d(\bm{x}_{i,\tau}) \leq \epsilon \\
		\infty & d(\bm{x}_{i,\tau}) < 0
	\end{array} \right.
\end{equation}
which increases when the signed distance $d(\bm{x}_{i,\tau})$ between $\mathcal{B}_i$ and its closest obstacle decreases and uses $\epsilon$ to define a buffer zone between the safe and unsafe areas. Then \eqref{eq:collisionField} accumulates \eqref{eq:collisionCost} in the time and geometry scales to generate the collision fields of \eqref{eq:forwardPF} and \eqref{eq:backwardPF}: 
\begin{equation} \label{eq:collisionField}
	c(\theta) = \sum_{\tau \in \mathcal{T}} \sum_{\mathcal{B}_i \in \bm{\mathcal{B}}} \alpha(t-\tau)\cdot c(\bm{x}_{i,\tau}), 
\end{equation}
where $\bm{\mathcal{B}} = \{\mathcal{B}_1, \dots,\mathcal{B}_{|\bm{\mathcal{B}}|}\}$ consists of all collision-check balls of a robot, and $\mathcal{T} = (t_1,t_2]$ and $\alpha$ are a time interval and a convolutional kernel respectively to ensure the continuous-time safety among the contiguous steps. \Cref{sec:parameters} will detail the choose of proper $\mathcal{T}$ and $\alpha$ facing the forward and backward planning in various scenarios. 

\subsection{Joint limit check}
Besides the collision avoidance, we also need to bound the joint state inside the available range to ensure the motion safety.  So we define a function   
\begin{equation} \label{eq:jointLimits}
	b(\bm{\theta}) = \left\{
	\begin{array}{cl}
		0  & \Sigma_{\text{d} = 1}^{D} |\theta_\text{d}^{\min} - \theta_\text{d}|^{+} + | \theta_\text{d} - \theta_\text{d}^{\max}|^{+}  = 0\\
		\infty & \text{Otherwise} \\
	\end{array}
	\right.
\end{equation}
to check whether a robot with $D$ degrees of freedom moves within the joint limits $[\bm{\theta}_{\min}, \bm{\theta}_{\max}]$. 

\section{METHODOLOGY}

\subsection{Gaussian mixture model for importance sampling} 
\Cref{eq:forwardPF,eq:backwardPF} indicate that the non-convexity mainly emerges from the collision repulsive field determined by the allocation of the obstacles and that the nonlinearity mainly originates from the forward kinematics used to check the collision and goal constraints. So it is impractical to directly sample from the so called true probabilistic distribution informed by the objective
\begin{equation} \label{eq:truePDF}
	p(\bm{\theta}|\mathcal{F}) = \frac{e^{-\varrho \mathcal{F}(\bm{\theta})}}{\int e^{-\varrho \mathcal{F}(\bm{\theta})} \diff \bm{\theta}}, 
\end{equation}
where $\varrho$ adjusts the objectives' effect on the true distribution and $\mathcal{F}_{goal}$ and $\mathcal{F}_{start}$ are both denoted by $\mathcal{F}$ due to our bidirectional method for their same online learning processes in the next section. 
%
%
Considering this situation, we adopt the Gaussian mixture model (GMM) used by \cite{}
\begin{equation} \label{eq:GMM}
	\hat{p}(\bm{\theta}|\mathcal{F}) = p(\bm{\theta}|\bm{\pi}, \bm{\mu}, \bm{\Sigma}) = \sum_{m = 1}^{M} \pi_m \mathcal{N}(\bm{\theta} | \bm{\mu}_m, \bm{\Sigma}_m), 
\end{equation}
where the mixture coefficient $\pi_m$ determines the importance weight of its corresponding modal and satisfies $\sum_{m = 1}^{M} \pi_m = 1$, to approximate $p(\bm{\theta}|\mathcal{F})$.

Unlike \cite{} using the GMM and autoregression methods for trajectory prediction, this paper uses GMM for Monte Carlo sampling and learns it online from the samples. Sequential importance sampling (SIS), as one of the popular Monte Carlo optimization methods, is adopted and the following section will detail how to learn the importance weights $\{\pi_m\}_{m = 1\dots M}$ (i.e., the mixture coefficients) sequentially and modify their corresponding model for a better approximation.

\subsection{Incrementally learned GMM for Bayes net construction} \label{sec:GMM}

\subsubsection{KL divergence} \label{sec:KL}
Since BN-MCO learns the GMM from the true posterior $p(\bm{\theta}|\mathcal{F})$ sequentially and samples new waypoints from it, it is essential to measure the divergence between the true and proposal posteriors. We deploy the KL-divergence to measure their cross-entropy: 
\begin{equation} \label{eq:KL-div}
\begin{aligned}
	D_{\text{KL}}(p || \hat{p}) =& \sum_{n = 1}^{N} p(\bm{\theta}_n|\mathcal{F}) \log \frac{p(\bm{\theta}_n|\mathcal{F})}{\hat{p}(\bm{\theta}_n|\mathcal{F})}\\
	=& \sum_{n = 1}^{N} p(\bm{\theta}_n|\mathcal{F}) \log \frac{p(\bm{\theta}_n|\mathcal{F})}{\sum_{m = 1}^{M} \pi_m \mathcal{N}(\bm{\theta}_n | \bm{\mu}_m, \bm{\Sigma}_m)}. 
\end{aligned}
\end{equation}

Then we estimate the likelihood of a sample $\bm{\theta}_n$ informed by the prior $\mathcal{N}(\bm{\theta}_n | \bm{\mu}_m^{(i)}, \bm{\Sigma}_m^{(i)})$ through $\nabla_{\bm{\pi}} D_{\text{KL}}(p || \hat{p})$:
\begin{align} \label{eq:singleImportance}
	\gamma_{n,m} &= \frac{p(\bm{\theta}_n|\mathcal{F}) \mathcal{N}(\bm{\theta}_n | \bm{\mu}_m^{(i)}, \bm{\Sigma}_m^{(i)})}{\sum_{m=1}^{M} \pi_{m}^{(i)} \mathcal{N}(\bm{\theta}_n | \bm{\mu}_m^{(i)}, \bm{\Sigma}_m^{(i)})}, 
\end{align}
and use it to get the estimated importances of each prior Gaussian distribution $\mathcal{N}^{(i)}_{m} = \mathcal{N}(\bm{\theta}|\bm{\mu}_m^{(i)},\bm{\Sigma}_m^{(i)})$:  
\begin{align} \label{eq:importance}	
	\pi_{m}^\text{est} &= \frac{\sum_{n = 1}^{N} \gamma_{n,m}}{\sum_{m = 1}^{M} \sum_{n = 1}^{N} \gamma_{n,m}}. 
\end{align}
Since the likelihood gathers the information of the true $p(\bm{\theta}|\mathcal{F})$ and prior $\mathcal{N}(\bm{\theta}|\bm{\mu},\bm{\mathcal{K}})$ distributions, $\argmin_{\bm{\theta}} \log \gamma$ is basically equal to the objective used by the numerical PF methods \eqref{eq:forwardPF-convex}-\eqref{eq:backwardPF-convex}. In this way, the factor $\eta_\pi \in (0,1)$ is used for incremental learning
\begin{align} \label{eq:importanceUpdate}
	\pi_{m}^{} & = \eta_\pi \pi_{m}^\text{est} + (1 - \eta_\pi)\pi_{m}^{(i)}, 
\end{align}
which gathers the historical information to prevent from the overfitting of GMM given a small dataset.  Then \Cref{alg:BayesNet} uses the updated importance $\pi_m^{}$ to renew 
\begin{equation} \label{eq:likelihoods}
	\gamma_{n,m} \leftarrow \frac{\gamma_{n,m}}{\sum_{n = 1}^{N} \gamma_{n,m}} \pi_m^{}
\end{equation}
and $\bm{\Theta}.\code{delete}(\{\gamma_{n,m}|^{m = 1\dots M}_{n = 1\dots N}\} <> 0)$ to delete the waypoints $\bm{\theta}_n \in \bm{\Theta}$ with $\gamma_{n,m} = 0$. 

\begin{algorithm}[htbp]
\caption{expBayesNet}\label{alg:BayesNet}
\DontPrintSemicolon
\LinesNumbered
\SetKwInOut{Input}{Input}
\SetKwInOut{Output}{Output}
\SetKwFunction{Union}{Union}
\Input {PF $\mathcal{F}$, waypoint set $\bm{\Theta}$, sampling number $N$.} 
\Output {Bayes Net $\{\bm{\mathcal{N}}, \bm{\mathcal{E}} \}$. }
\textbf{Initialization:} initial effector $\varrho_0$ and its adjustor $\eta_\varrho$, learning factors $\{\eta_\pi, \eta_\mu, \eta_\sigma\}$, initial $\hat{p}^{(1)}(\bm{\theta}|\mathcal{F})$.\; 

\For(\tcp*[h]{\scriptsize expand Bayes net via Monte Carlo optimization}){$ \textit{MCOIter}: i = 1 \dots N_\text{mco} $}{
	Sample a waypoint set $\bm{\Theta}$ from a GMM $\hat{p}^{(i)}$;\;
	Update the importances 
	$\bm{\pi}$ 
	of GMM by \eqref{eq:importanceUpdate}; \;
	Renew the likelihood $\gamma$ by \eqref{eq:likelihoods}; \;
	Divide $\bm{\Theta}$ into $L$ clusters: $\bigcup_{l = 1}^{L} \bm{\Theta}_l \subseteq \bm{\Theta}$; \;
	\For(\tcp*[h]{\footnotesize cluster analysis of sampled waypoints}){$\textit{clusterIter}: l = 1 \dots L$}{
		Find all models $\{\mathcal{N}_{m}^{(i)}|_{m = 1\dots M_l}\}$ that generate the cluster $\bm{\Theta}_l$; \;
		\For(\tcp*[h]{\footnotesize edge stretches from prior to new nodes}){$ \textit{stretchIter}: m = 1 \dots M_l $}{
			Learn a new node $\mathcal{N}_{lm}^{}$ incrementally from $\bm{\Theta}_{l}$ based on $\mathcal{N}_{m}^{(i)}$ by \eqref{eq:iGMM}; \;
			Stretch an edge $\mathcal{E}^{}.\code{dir} = \{m,lm\}$ from $\mathcal{N}_{m}^{(i)}$ to $\mathcal{N}_{lm}^{}$ with importance $\pi_{lm}^{}$ and sampling number $N_{lm}^{}$ by \eqref{eq:edge}-\eqref{eq:weight}; \;
			\If {$N_{lm}^{} > \mathcal{E}\text{tol}$}{
				Cache the info of $\mathcal{N}^{}_{lm},\mathcal{E},\pi^{}_{lm},N_{lm}^{}$ in $\bm{\mathcal{N}}, \bm{\mathcal{E}}, \bm{\pi}^{(i+1)}, \bm{N}$, respectively; \;
				\lIf{$\mathcal{N}^{}_{lm}$ satisfies the constraint $\mathcal{F}$}{
					\Return. 
				}
			}
		}
	}
	Gain a new GMM $\hat{p}^{(i+1)}$ and adjust PF by \eqref{eq:rho}; \; 
}
\end{algorithm}
%

\subsubsection{Cluster} \label{sec:cluster}
Due to the high dimensionality and nonlinearity of $\log p(\bm{\theta}|\mathcal{F})$ even introducing the convex part $\mathcal{K}$, this paper utilizes cluster analysis to find its features and learn the GMM incrementally from them. For computational simplification, we directly use a symmetric matrix $\mathbf{D} = [d_{ij}]_{i,j  = 1\dots N}$ to store the euclidean distances between any two samples $\bm{\theta}_i, \bm{\theta}_j$. Then we find the \textit{k}-nearest neighbors of each sample $\bm{\theta}_i$, set their corresponding columns as 1 and the others as 0 in row $i$, and gain a binary matrix $\mathbf{N}$, each of whose rows stores its nearest neighbors. Finally, we gain a matrix $\mathbf{N}' = \mathbf{N} + \mathbf{N}^\text{T} > 2$ whose none zero elements $n'_{ij},n'_{ji} = 1$ indicate samples $\bm{\theta}_i$ and $\bm{\theta}_j$ are one of the $k$-nearest neighbors of each other, a.k.a. related. In this way, the related samples construct a cluster $\bm{\Theta}_{l}$, where $\forall \bm{\theta}_i, \bm{\theta}_j \in \bm{\Theta}_{l} \text{ w/ } n'_{ij} = 1$, $\bigcup_{l = 1}^{L} \bm{\Theta}_{l} \subseteq \bm{\Theta}$ and $\bm{\Theta}_{l_1} \bigcap \bm{\Theta}_{l_2} = \emptyset$. Different from the former methods~\cite{} generating the separated samples hierarchically for a preset number of clusters, we set the number of nearest neighbor \textit{k} instead and get the clusters.

\subsubsection{Node} \label{sec:node}
Since the sampled waypoints of a cluster $\bm{\Theta}_k$ may originate from more than one Gaussian models, we first estimate the models informed by $\bm{\Theta}_k$ through solving $\argmin_{\bm{\mu},\bm{\mathcal{K}}} D_{\text{KL}}(p || \hat{p})$ and get
\begin{equation} \label{eq:estGMM}
\begin{gathered}
	\bm{\mu}_{l,m}^\text{est} = \frac{\sum_{\bm{\theta}_n \in \bm{\Theta}_l } \gamma_{n,m} \bm{\theta}_n}{\sum_{\bm{\theta}_n \in \bm{\Theta}_l } \gamma_{n,m} }, \\
%
%
	\bm{\Sigma}_{l,m}^\text{est} = \frac{\sum_{\bm{\theta}_n \in \bm{\Theta}_l } \gamma_{n,m} (\bm{\theta}_n  - \bm{\mu}_m^{(i)}) (\bm{\theta}_n  - \bm{\mu}_m^{(i)})^{\text{T}} }{\sum_{\bm{\theta}_n \in \bm{\Theta}_l } \gamma_{n,m} }, 
\end{gathered}
\end{equation}
where $\forall m \in \{1\dots M_l\}$, an index set of the models generating $\bm{\Theta}_l \ni \bm{\theta}_n$, satisfies $\exists \bm{\theta}_n \backsim \mathcal{N}^{(i)}_m$. Then the factors $\eta_\mu, \eta_\sigma \in (0,1)$ proceed the incremental online learning of $\mathcal{N}^{(i)}_{m}$ to renew the GMM and generate new nodes: 
\begin{equation} \label{eq:iGMM}
\begin{aligned}
	\bm{\mu}_{l,m}^{} &= \eta_\mu \bm{\mu}_m^{(i)} + (1-\eta_\mu) \bm{\mu}_{l,m}^\text{est}, \\
	\bm{\Sigma}_{l,m}^{} &= \eta_\sigma \bm{\Sigma}_m^{(i)} + (1-\eta_\sigma) \bm{\Sigma}_{l,m}^\text{est}. 
\end{aligned}
\end{equation}

\begin{figure}[h]
 	\centering
	\includegraphics[width=1\linewidth]{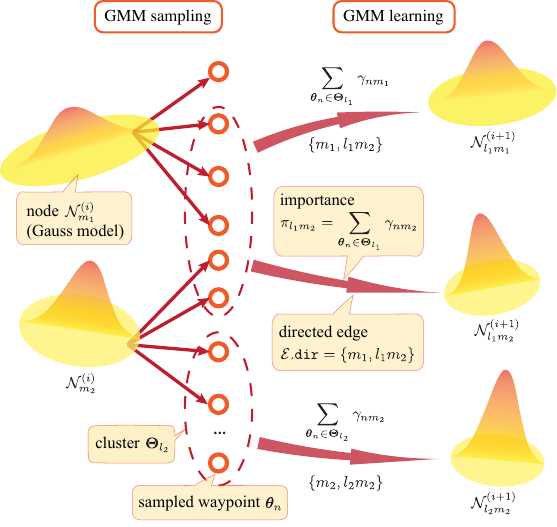}
	\caption{ The waypoints $\bm{\theta}_n$, sampled from the nodes $\mathcal{N}_{m_1}^{(i)}, \mathcal{N}_{m_2}^{(i)}$ in the $i^\text{th}$ layer of Bayes net, are divided into clusters $\bm{\Theta}_{l_1}, \bm{\Theta}_{l_2}$ according to \Cref{sec:cluster}. Nodes $\mathcal{N}_{m_1}^{(i)},\mathcal{N}_{m_2}^{(i)}$ stretch out edges $\{m_1, l_1m_1\}, \{m_2, l_1 m_2\}$ to generate new nodes $\mathcal{N}_{l_1 m_1}^{(i+1)}, \mathcal{N}_{l_1 m_2}^{(i+1)}$, while node $\mathcal{N}_{m_2}^{(i)}$ stretches out edge $\{m_2, l_2 m_2\}$ to generate a new node $\mathcal{N}_{l_2 m_2}^{(i+1)}$ through the incremental GMM learning (\Cref{sec:node}). 
	\label{fig:expBayesNet}}
\end{figure}

\Cref{fig:expBayesNet} shows how a node $\mathcal{N}^{(i)}_{m_1}$ stretches out two new nodes, a.k.a. models, $\mathcal{N}^{(i+1)}_{l_1 m_1}, \mathcal{N}^{(i+1)}_{l_2 m_1}$ measured by two posteriors $p(\bm{\Theta}_{l_1}|\mathcal{F}), p(\bm{\Theta}_{l_2}|\mathcal{F})$ respectively, and how the same posterior $p(\bm{\Theta}_{l_1}|\mathcal{F})$ informs another prior model $\mathcal{N}_{m_2}^{(i)}$ to get a new node $\mathcal{N}_{l_1 m_2}^{(i+1)}$. 

%
%

\subsubsection{Edge} \label{sec:edge}
During the Bayes net expansion with node stretching, the directed edge $\mathcal{E}.\code{dir} = (m_1,m_2)^{}$ connects from a prior model $\mathcal{N}_{m_1}^{(i)}$ in the $i^\text{th}$ layer to another node $\mathcal{N}_{m_2}^{} = \mathcal{N}_{l_1 m_1}^{}$ updated from the prior and each layer denotes a single procedure of GMM's incremental learning. Expect for the connectivity and direction, the weight is another important property of the edge $\mathcal{E}$ and the negative logarithm is adopted for the importance calculation: 
\begin{equation} \label{eq:edge}
\begin{array}{c}
	\mathcal{E}.\code{length} = - \log \left( \sum_{\bm{\theta}_n^{} \in \bm{\Theta}_{l_1}^{(i)}} \gamma_{n,m_1}^{} \right). 
\end{array}
\end{equation}
Not only does it contain the Mahalanobis distance $(\|\bm{\theta}_n - \bm{\mu}_{m_1}^{(i)}\|_{(\bm{\Sigma}_{m_1}^{(i)})^{-1}}^{2})^{-\frac{1}{2}}$ between the waypoints contained in cluster $\bm{\Theta}^{(i)}_{l_1}$ and the expectation value of their common prior $\mathcal{N}_{m_1}^{(i)}$, but indicate how $\bm{\theta}_n$ satisfies the constraint $\mathcal{F}$ according to the  definition of $\gamma_{n,m_1}$ in \eqref{eq:singleImportance}. Moreover, \Cref{alg:BayesNet} uses the above information to update the importance and its corresponding sampling number of each model: 
\begin{equation} \label{eq:weight}
	\pi_{m_2}^{} = \exp(-\mathcal{E}.\code{length}),~
	N_{m_2}^{} = N \cdot \pi_{m_2}, 
\end{equation}
because the new waypoints are sampled from the GMM~\eqref{eq:GMM}. 

Up to now, we have introduced all the technologies of Bayes net construction during the incremental learning and Monte Carlo optimization. Since their stretching process introduces redundant edges, complicating the net structure, we use a factor $\mathcal{E}\text{tol}$ to refine the process: if $N_{m_2}  > \mathcal{E}\text{tol}$, meaning that the node $\mathcal{N}_{m_2}$ is qualified, it will be added to the node set $\bm{\mathcal{N}}^{}$, and the sets ${\bm{\mathcal{E}}^{}, \bm{\pi}^{(i+1)}, \bm{N}}$ will cache the information of edge\footnote{
	Since $\mathcal{E}.\code{dir} = (m_1,m_2)$ stores the indexes of the connected nodes $\mathcal{N}_{m_1},\mathcal{N}_{m_2}$, its cache process renews the index information additionally according to the realigned order of $\mathcal{N}_{m_1},\mathcal{N}_{m_2}$. 
} $\mathcal{E}$,  importance $\pi_{m_2}$ and sampling number $N_{m_2}$; 
otherwise, $\mathcal{N}_{m_2}$ will be rejected. At the end of each learning process (\textit{MCOIter}), \Cref{alg:BayesNet} will gain a new GMM $\hat{p}^{(i+1)}$ informed by the weighted combination of nodes/models from $\hat{p}^{(i)}$ due to \eqref{eq:GMM} and update the factor  
\begin{equation} \label{eq:rho}
	\varrho = \varrho_0 (i+1)^{\eta_\varrho}
\end{equation}
of the true distribution \eqref{eq:truePDF} to adjust the effects of the PFs \eqref{eq:forwardPF}-\eqref{eq:backwardPF} sequentially.

\subsection{Path finding by bidirectional Bayes net} \label{sec:findPath}

The above section have introduced how the bidirectional Bayes nets $\{\bm{\mathcal{N}}_{s}, \bm{\mathcal{E}}_{s}\}, \{\bm{\mathcal{N}}_{g}, \bm{\mathcal{E}}_{g}\}$, informed by $\mathcal{F}_{start}$~\eqref{eq:forwardPF} and $\mathcal{F}_{goal}$~\eqref{eq:backwardPF}, correspondingly, are constructed through \code{expBayesNet} (\Cref{alg:BayesNet}) to approach the null space of the constraints. This section will introduce how \code{BN-MCO} (\Cref{alg:BN-MCO}) finds a feasible path through an all pair shortest path algorithm \code{Dijkstra}. 

To ensure the connectivity between the bidirectional Bayes nets, BN-MCO first samples a set of $N$ waypoints $\bm{\Theta}$ uniformly from the space restricted by $[\bm{\theta}_{\min}, \bm{\theta}_{\max}]$ and estimates $\bm{\Theta}$'s likelihoods 
\begin{align} \label{eq:singleImportance_1}
	\gamma_{n} &= \frac{p(\bm{\theta}_n|\mathcal{F})}{\sum_{n=1}^{N}p(\bm{\theta}_n|\mathcal{F})} , 
\end{align}
to filter out the useless samples $\bm{\Theta}.\code{delete}(\gamma_{n} <> 0)$. Different from \eqref{eq:singleImportance} take the priors and posteriors into consideration, the above equation only uses the posteriors because the prior information of each sample is uniform. BN-MCO then constructs the Bayes nets of $\mathcal{F}_{goal}$ and $\mathcal{F}_{start}$ separately from the same waypoint set $\bm{\Theta}_0$. A roadmap object $\bm{\mathcal{M}}$, constructed by \code{roadMapConstruct}, originates from a node $\mathcal{N}^*$ satisfying a constraint $\mathcal{F}$ and stores all the nodes connecting with each others as well as the corresponding edges related in order 
under the connectivity test. 
Each element of the matrix $\bm{\mathcal{M}}_{ij}$ in row-$i$ and column-$j$ stores the weight of edge that connects \code{$\bm{\mathcal{M}}$.node($i$)} and \code{$\bm{\mathcal{M}}$.node($j$)}.  

\begin{algorithm}[htbp]
\caption{BN-MCO}\label{alg:BN-MCO}
\DontPrintSemicolon
\LinesNumbered
\SetKw{and}{and}
\SetKwInOut{Input}{Input}
\SetKwInOut{Output}{Output}
\SetKwFunction{RPC}{roadMapConstruct}
\SetKwFunction{RM}{$\bm{\mathcal{M}}$}
\SetKwFunction{addnode}{addNode}
\SetKwFunction{addedge}{addEdge}
\SetKwFunction{isconnect}{isConnect}
\SetKwFunction{node}{node}
\SetKwFunction{edge}{edge}
\SetKwFunction{findconnect}{findConnect}
\SetKwFunction{SP}{Dijkstra}
\SetKwFunction{findtraj}{findTrajectory}
\SetKwProg{Fn}{def}{\string:}{}
\Input {forward PF $\mathcal{F}_{goal}$, backward PF $\mathcal{F}_{start}$.}
\Output {path $\bm{\xi}^*$. }
$\bm{\Theta}_0  = \{\bm{\theta}_n \backsim \mathcal{U}(\bm{\theta}_{\min},\bm{\theta}_{\max}) |_{n = 1\dots N} \}$; \tcp*[h]{\scriptsize sample waypoints from uniform distribution} \;
Calculate likelihoods $\{\gamma_{n}|_{n = 1\dots N}\}$ by \eqref{eq:singleImportance_1}; \;
$\bm{\Theta}_0.\code{delete($\{\gamma_{n}\} <> 0$)}$; \;

$\{\bm{\mathcal{N}}_{g}, \bm{\mathcal{E}}_{g}\} = \code{expBayesNet($\mathcal{F}_{goal}$,$\bm{\Theta}_0$,$N$)}$; \;
$\{\bm{\mathcal{N}}_{s}, \bm{\mathcal{E}}_{s}\} = \code{expBayesNet($\mathcal{F}_{start}$,$\bm{\Theta}_0$,$N$)}$; \;
$\RM_g = $ \RPC{$\bm{\mathcal{N}}_g, \bm{\mathcal{E}}_g, \mathcal{F}_{goal}$}; \;
$\RM_s = $ \RPC{$\bm{\mathcal{N}}_s, \bm{\mathcal{E}}_s, \mathcal{F}_{start}$}; \;
Find all pair shortest paths from the nodes of $\RM_g, \RM_s$ that satisfy the constraints by $\bm{\Xi}_g$ = \SP{$\RM_g,1$}, $\bm{\Xi}_s$ = \SP{$\RM_s,1$}; \;

\tcc*[h]{\scriptsize connect the paths of $\bm{\Xi}_s,\bm{\Xi}_g$ sharing nodes}\;
Connect all pairs $\bm{\mathcal{E}}^\text{conn}$ of the nodes from $\RM_g.\node$ to $\RM_s.\node$ and initialize a set of paths $\bm{\Xi} = \emptyset$; \; 
\For{$i = 1\dots |\bm{\mathcal{E}}^\text{conn}|$}{
	$\bm{\Xi} \leftarrow \bm{\Xi} \bigcup \bm{\xi}_i $ with a pair $\bm{\mathcal{E}}^\text{conn}_{i}$ by \eqref{eq:path}-\eqref{eq:pathLength}; \; 
	$\bm{\Xi}.\code{sort}$; \tcp*[h]{\scriptsize in ascending order due to $\bm{\xi}.\code{length}$} \; 
}
\For{$i = 1\dots |\bm{\Xi}|$}{
	$\bm{\xi}^* = \bm{\xi}_i$.\findtraj; \;
	\lIf{$\bm{\xi}^* \notin \emptyset$}{
		\Return $\bm{\xi}^*$. 
	}
}


\Fn{\findtraj{$\bm{\xi}$, $\bm{\theta}$}}{
	\lIf{$\bm{\theta} \in \emptyset$}{
		$\bm{\theta} = \bm{\xi}.\code{init}$;  
		$\bm{\xi}.\code{delete($\bm{\theta}$)}$; 
	}
	$\bm{\xi}^*.\code{add($\bm{\theta}$)}$; 
	$\bm{\Theta}_\text{safe} = \bm{\xi}.\code{findSafePoints($\bm{\theta}$)}$;\;
	\lIf{$\bm{\theta} \in \bm{\xi}.\code{end}$}{
		\Return $\bm{\xi}^*$. 
	}
	\lFor{$\forall \bm{\theta} \in \bm{\Theta}_\text{safe}$}{
		$\bm{\xi}$.\findtraj{$\bm{\theta}$};
	}
}

\end{algorithm}
 \begin{figure}[h]
 	\includegraphics[width=1\linewidth]{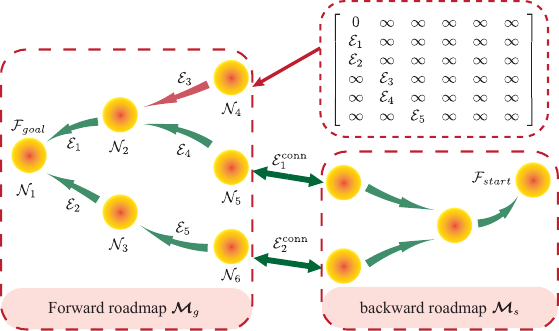}
	\caption{
	The forward and backward roadmaps $\bm{\mathcal{M}}_g$ and $\bm{\mathcal{M}}_s$, constructed by \code{expBayesNet} (\Cref{alg:BayesNet}) in bi-direction, connects the Gaussian models denoted by yellow nodes sharing the same waypoints to pair with each other via $\mathcal{E}^\text{conn}$ and finds the paths from the initial to goal states. 
	}
	\label{fig:pairNet}
\end{figure}
\begin{figure}[h]	
	\includegraphics[width=1\linewidth]{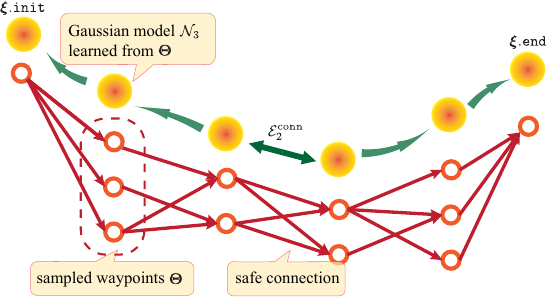}
	\caption{
	A path $\bm{\xi}$ with $\mathcal{E}^\text{conn}_2$ extracted from the paired map of \Cref{fig:pairNet} consists of serially connected yellow nodes (i.e., Gaussian models), each of whom contains the sampled waypoints gathered for its incremental learning. $\bm{\xi}.\code{findTrajectory}$ of \Cref{alg:BN-MCO} builds the safe connections among the waypoints of the adjacent nodes recursively from $\bm{\xi}.\code{init}$ to $\bm{\xi}.\code{end}$ to find a time-continuous safe trajectory.  
	}
	\label{fig:findTraj}
\end{figure}

After that, Dijkstra~\cite{Dijkstra1959Dijkstra-Alg}, an all pair shortest path algorithm, searches for the shortest paths from the origin of $\bm{\mathcal{M}}$ and stores the information of the paths and their lengths in \code{$\bm{\Xi}$.path} and \code{$\bm{\Xi}$.length}. Considering the separation of the bidirectional roadmaps ($\bm{\mathcal{M}}_g$ and $\bm{\mathcal{M}}_s$), BN-MCO pairs up their nodes which share the same subset of $\bm{\Theta}_0$ sampled uniformly at the beginning in an object $\bm{\mathcal{E}}^\text{conn}$, where $\bm{\mathcal{E}}^\text{conn}_i$ pairs a node $\bm{\mathcal{E}}^\text{conn}_i.\code{pair(1)}$ from $\bm{\mathcal{M}}_s$ with $\bm{\mathcal{E}}^\text{conn}_i.\code{pair(2)}$ from $\bm{\mathcal{M}}_g$. \Cref{fig:pairNet} visualizes the pairing process and how the two roadmaps constructed in opposite direction connect with each other by $\bm{\mathcal{E}}^\text{conn}_i$ to find a feasible path 
\begin{equation} \label{eq:path}
\begin{aligned}
	\bm{\xi}_i.\code{path} =& \{ \bm{\Xi}_s.\code{path}(\bm{\mathcal{E}}^\text{conn}_{i}.\code{pair(1)}).\code{reverse}, \\
	& \bm{\mathcal{E}}^\text{conn}_{i}.\code{pair}, \bm{\Xi}_g.\code{path}(\bm{\mathcal{E}}^\text{conn}_{i}.\code{pair(2)})\} 
\end{aligned}
\end{equation}
from the start to the goal constraint via the pair $\bm{\mathcal{E}}^\text{conn}_{i}.\code{pair}$. Moreover, we also calculate the length of $\bm{\xi}_i$ by
\begin{equation}\label{eq:pathLength}
\begin{aligned}
	\bm{\xi}_i.\code{length} &= \bm{\Xi}_s.\code{length}(\bm{\mathcal{E}}^\text{conn}_{i}.\code{pair(1)}) \\
	& + \bm{\mathcal{E}}^\text{conn}_{i}.\code{length} \\
	& + \bm{\Xi}_g.\code{length}(\bm{\mathcal{E}}^\text{conn}_{i}.\code{pair(2)}), 
\end{aligned}
\end{equation}
store all paired paths in $\bm{\Xi}$, and sort them in an ascending order due to their length. 
Since the length/weight of edge \code{$\mathcal{E}$.length} depends on the logarithm of likelihood due to \eqref{eq:singleImportance} and \eqref{eq:edge}, a smaller path length basically means a higher probability of the Markov chain: 
\begin{equation}
\begin{aligned}
	& \min \bm{\xi}.\code{length} = \sum_{j=1}^{|\bm{\xi}|} \mathcal{E}^{(j,j-1)} \Leftrightarrow \\
	& \max \prod_{j = 1}^{|\bm{\xi}|} \hat{p}(\bm{\theta}^{(j)}|\bm{\theta}^{(j-1)}, \mathcal{F}) \cdot p(\bm{\theta}^{(j)}|\mathcal{F}). 
\end{aligned} 
\end{equation}

However, the Markov chain only describes the probabilistic transition from the initial state to the states satisfying the constraint $\mathcal{F}$ rather than the actual state transition, especially when a node $\mathcal{N}$ has a large covariance $\bm{\Sigma}$, meaning there exists a large bias between its contained samples and its expectation $\bm{\mu}$. In this way, \code{findTrajectory($\bm{\xi}$,$\bm{\theta}$)}, a recursive method of a path object $\bm{\xi}$, tends to find a trajectory of the transition. As shown in \Cref{fig:findTraj}, it starts from an initial node $\bm{\theta} = \bm{\xi}.\code{init}$ and finds the waypoints $\bm{\Theta}_\text{safe}$ from its connected node\footnote{Since each node of the Bayes net are generated through the incremental learning from the waypoints sampled from the previous GMM due to \Cref{alg:BayesNet}, each node stores the information of waypoints used for incremental learning. }, all of which connect safely to $\bm{\xi}.\code{init}$, through $\bm{\xi}.\code{findSafePoints($\bm{\theta}$)}$. The safe connection ensures not only each waypoint collision-free but the time continuous safety between any two adjacent waypoints. Then each safely connected waypoint will continue finding their $\bm{\Theta}_\text{safe}$ and utilize them to construct trajectory recursively until $\bm{\Theta}_\text{safe} = \emptyset$ or $\bm{\theta}$ reaches the end of a path $\bm{\xi}.\code{end}$.

\section{EXPERIMENT}

\subsection{Evaluation} \label{sec:Evaluation}

\subsubsection{Setup for benchmark}\label{sec:setup}

Since this paper focuses on the searching efficiency of the sampling based motion planners and utilizes  the KL-divergence to find a path in a numerical way, we benchmarks BN-MCO against the numerical (TrajOpt~\cite{Schulman2013SCO}, GPMP2~\cite{Mukadam2018GPMP}) and sampling  methods (PRM~\cite{Kavraki1994PRM}, RRT-Connect~\cite{Kuffner2000RRT-connect}, RRT*~\cite{Karaman2011RRT*-PRM*}).  The benchmark proceeds on MATLAB with the encoded kernel algorithm in C++. 

\begin{figure}[h]
	\begin{centering}
		\begin{subfigure}[b]{0.238\textwidth}
			\centering
			\includegraphics[width=1\linewidth]{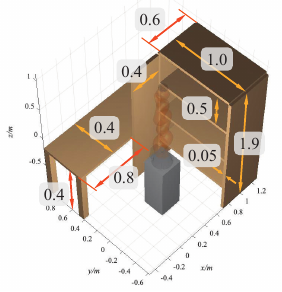} 
			\caption{Planning scenario}
			\label{fig:desktop}
		\end{subfigure}
		\hfill
		\begin{subfigure}[b]{0.238\textwidth}
			\centering
			\includegraphics[width=1\linewidth]{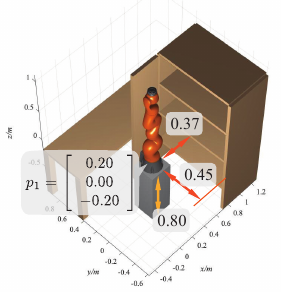}
			\caption{Stand position 1, $p_1$}
			\label{fig:p1}
		\end{subfigure}
		\\
		\begin{subfigure}[b]{0.238\textwidth}
			\centering
			\includegraphics[width=1\linewidth]{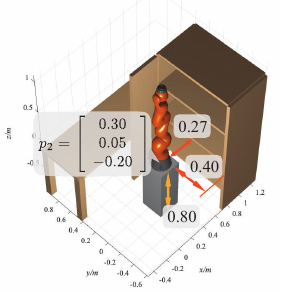}
			\caption{Stand position 2, $p_2$}
			\label{fig:p2}
		\end{subfigure}
		\hfill
		\begin{subfigure}[b]{0.238\textwidth}
			\centering
			\includegraphics[width=1\linewidth]{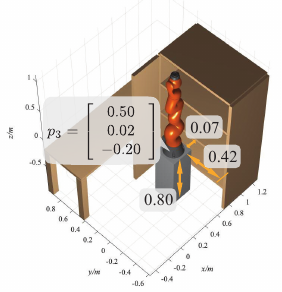}
			\caption{Stand position 3, $p_3$}
			\label{fig:p3}
		\end{subfigure}
	\end{centering}
	\caption{The initial trajectory with red collision parts visualizes our benchmark on LBR-iiwa or AUBO-i5. \textit{Task C-6} means the No.{6} task of class {C}. 
	\label{fig:problems}}
\end{figure}

To illustrate the competence of BN-MCO for planning tasks, we conduct 36 experiments on LBR-iiwa, a manipulator with 7 degrees of freedom, at a bookshelf attached with a desktop. \Cref{fig:problems} shows how we place the manipulator on a stand and change its location as well as that of the cup it tends to grasp to realize variant planning task. Since the height of the stand does have a significant effect on the reachability of the manipulator but relative distance between the manipulator and the bookshelf does, our experiment keeps the height $z_s$ constant and adjusts the planar parameter $x_s$ and $y_s$, as shown in \Cref{tab:standPosition}. As for the goal constraint, \Cref{tab:goalConstraint} utilizes $[x, y, z,\psi, \theta, \phi]$ to denote $[x_1, \dots, x_6]$ of \eqref{eq:goalConstraint} and sets 5 different constraints $[g_1, \dots, g_5]$, only setting an available range of $\phi = x_6$ with $[\phi_{\min},\phi_{\max}]$ and the other parameters with a specific value. 
Meanwhile, \Cref{tab:startConstraint} sets 3 different start constraints $[s_1, s_2, s_3]$. 
Then \Cref{tab:taskSet} sets up 36 different tasks 
%
	$\mathcal{T} = \{g, s, p\}$
%
in 3 dimensions ($\mathcal{F}_{goal}$, $\mathcal{F}_{start}$, $p_\text{stand}$) based on \Cref{tab:standPosition,tab:goalConstraint,tab:startConstraint}. 

\begin{table}[htbp]
\label{table:task}
\centering
\begin{threeparttable}
\begin{subtable}{.5\textwidth}
\centering
\caption{Stand position $p_\text{stand} = [x_s, y_s, z_s]$} \label{tab:standPosition}
\scalebox{1}{
\begin{tabular}{ccccccc}
\toprule
$p_{1}$ & $p_{2}$ & $p_{3}$  \\
\midrule
$[0.20, 0.00, -0.20]$ & $[0.30, 0.05, -0.20]$ & $[0.50, 0.02, -0.20]$ \\
\bottomrule
\end{tabular}}
\end{subtable}
\par \bigskip
\begin{subtable}[h]{.5\textwidth}
\centering
\caption{Goal constraints $\mathcal{F}_{goal}$} \label{tab:goalConstraint}
\scalebox{1}{
\begin{tabular}{llcccccccccccccccc}
\toprule
& $x$ & $y$ & $z$ & $\psi$ & $\theta$ & $\phi_{\min}$ & $\phi_{\max}$ \\ 
\midrule
$g_1$ & 0.0 & 0.6 & -0.1 & 0.0 & $-{\pi}/{2}$ & $0.0$ & $\pi$\\
$g_2$ & 0.2 & 0.6 & -0.1 & 0.0 & $-{\pi}/{2}$ & $0.0$ & $\pi$\\
$g_3$ & 1.1 & 0.0 & 0.5 & 0.0 & $-{\pi}/{2}$ & $-\pi/2$ & $\pi/2$ \\
$g_4$ & 0.9 &  0.0 & 0.0 & 0.0 & $-{\pi}/{2}$ & $-\pi/2$ & $\pi/2$ \\
$g_5$ & 1.2 & 0.0 & 0.0 & 0.0 & $-{\pi}/{2}$ & $-\pi/2$ & $\pi/2$ \\
\bottomrule
\end{tabular}}
 \begin{tablenotes}\scriptsize
      \item[1] The units of $x, y, z$ and $\psi, \theta, \phi$ are meter and radius, respectively. 
 \end{tablenotes}
\end{subtable}
\par \bigskip
\begin{subtable}{.5\textwidth}
\centering
\caption{Start constraints $\mathcal{F}_{start}$} \label{tab:startConstraint}
\scalebox{1}{
\begin{tabular}{lccccccc}
\toprule
$s_1$ & [0.00, 0.00, 0.00, 0.00, 0.00, 0.00, 0.00] \\
$s_2$ & [2.06, 1.73, 0.99, 0.80, 1.99, -0.41, -2.11] \\
$s_3$ & [-1.37, -0.29, -0.29, 0.85, 1.45, -0.56, -0.71] \\
\bottomrule
\end{tabular}}
\end{subtable}
\par \bigskip
\begin{subtable}{.5\textwidth}
\caption{Task setting} \label{tab:taskSet}
\centering
\scalebox{1}{
\begin{tabular}{cccccccc}
\toprule
\scalebox{0.75}{\diagbox{$\mathcal{F}_{start}$}{$p_\text{stand}$}{$\mathcal{F}_{goal}$}} & $g_1$ & $g_2$ & $g_3$ & $g_4$ & $g_5$\\ 
\midrule
 $s_1$ & $p_{1},p_{2}$ & $p_{1},p_{3}$ & $p_{1},p_{2},p_{3}$ & $p_{2},p_{3}$ & $p_{1},p_{2},p_{3}$ \\
 $s_2$ & $p_{1},p_{3}$ & $p_{1},p_{3}$ & $p_{1},p_{2},p_{3}$ & $p_{1},p_{3}$ & $p_{1},p_{2},p_{3}$  \\
 $s_3$ & $p_{2},p_{3}$ & $p_{1},p_{3}$ & $p_{1},p_{2},p_{3}$ & $p_{1},p_{2}$ & $p_{1},p_{2},p_{3}$  \\
\bottomrule
\end{tabular}}
\end{subtable}
\end{threeparttable}
\end{table}

 Besides, we introduce the $\beta$-distribution into the collision field calculation \eqref{eq:collisionField} for the time continuous safety. 

%

\subsubsection{Parameter setting}\label{sec:parameters}

The online learning is one of the most important parts of the Bayes net expansion, which informs the feasible paths finding. In this way, this section will first tune three key parameters $\{\eta_\pi, \eta_\mu, \eta_\sigma\}$ used for incremental learning and analyze how they affect BN-MCO's performance in 10 tasks $\{\mathcal{T}_{12}, \mathcal{T}_{14},\mathcal{T}_{16},\mathcal{T}_{19},\mathcal{T}_{21},\mathcal{T}_{24},\mathcal{T}_{26},\mathcal{T}_{28},\mathcal{T}_{31},\mathcal{T}_{36}\}$. Additionally, we also tune the factor $\beta$ and analyze its results, because the $\beta$-distribution determines how to select the internal time $\tau\in(t,t+1)$ between any two adjacent waypoints $\bm{\theta}_{t}, \bm{\theta}_{t+1}$. 

\begin{table}
\caption{Tuning results of $\eta_\pi, \eta_\mu, \eta_\delta$ in 10 tasks}
\label{tab:tuning_eta}
\centering
\begin{threeparttable}
{
\begin{tabular}{c|ccccccc}
\toprule
\scalebox{0.8}{\diagbox{$\eta_\pi$}{(\%)|(s)}{$\eta_\mu / \eta_\delta$}} & 0.2 / 0.1 & 0.4 / 0.2 & 0.8 / 0.4 & 1.0 / 0.5 \\ 
\midrule
 0.20 & 20 | 16.7 & 60 | 14.4 & 40 | 16.3 & 0 | 20 \\
 0.40 & \bf{80} | 7.03 & 50 | 11.8 & 30 | 13.8 & 10 | 16.2 \\
 0.80 & 60 | 9.25 & 70 | 9.94 & 50 | 12.2 & 10 | 17.3 \\
 1.00 & 30 | 14.8 & 20 | 16.4 & 20 | 17.1 & 0 | 20\\
\bottomrule
\end{tabular}}
\begin{tablenotes}\scriptsize
     \item[1] (\%) and (s) denote the success rate and average computation time. 
\end{tablenotes}
\end{threeparttable}
\end{table}

\Cref{tab:tuning_eta} verifies that the value of learning factors $\eta_\pi, \eta_\mu, \eta_\delta$ as well as the fraction among them have large effects on the planning. The small value $\eta_\pi = 0.2$ dampens the online learning, resulting in the low success rate (8\%-20\%) and high computational time (14s-20s). Meanwhile, the large value $\eta_\pi = 1.00$ leads to a rapid convergence to some local minima informed by the dataset gathered from the primary steps and has a success rate (0-30\%) and large computation time (14s-20s). So a proper value $\eta_\pi = 0.40$ can help gather more comprehensive information from the incrementally learned GMM and improve the success rate to 10\%-80\% and save the computation time to 7s-16s. Moreover, this table shows the success rate increases from 0-50\% to 20\%-80\% when the fraction $\eta_\pi / \eta_\mu$ approaches 2.0, meaning the asychronized learning of $\eta_\pi$ and $\eta_\mu, \eta_\sigma$ can enhance the leaning ability.  

\begin{table}
\caption{Tuning results of $\beta$ in 10 tasks}
\label{tab:tuning_beta}
\centering
\begin{threeparttable}
{
\begin{tabular}{cccccccc}
\toprule
$\beta$ & 0.1 & 0.2 & 0.5 & 1.0 & 2.0 \\ 
\midrule
(\%)|(s) & 40 | 11.2 & 80 | 7.03 & 70 | 9.34 & 20 | 17.9 & 30 | 14.8 \\
\bottomrule
\end{tabular}}
\end{threeparttable}
\end{table}

Since the distribution becomes uniform when $\beta = 1.0$, it divides the bunch of distributions into 
the convex ($\beta < 1.0 $) and concave ($\beta > 1.0$) parts. As a result, \Cref{tab:tuning_beta} shows that a small $\beta = 0.1$, meaning that the internal point nearby the adjacent points can be  selected with high probability under the convex distribution, results in the low success rate (40\%) due to the inefficient continuous check. Meanwhile, the large $\beta \geq 1.0$, meaning the selected internal points aggregate nearby the mid-point $\bm{\theta}_{t+\frac{1}{2}}$ under the concave distribution, also leads to the low success rate (20\%), especially confronting the narrow feasible workspace. 

\subsubsection{Result analysis}\label{sec:analysis}
\begin{table*}[hbtp]
\caption{Results of 36 tasks with 5 repeated tests}
\label{tab:results:iiwa}
\centering
\begin{threeparttable}[c]
\scalebox{1}{
\begin{tabular}{lccc|cc|ccccc}
\toprule
& \multicolumn{3}{c|}{Our method}  & \multicolumn{2}{c|}{numerical optimization} & \multicolumn{3}{c}{probabilistic sampling} \\
& \bf{BN-MCO-400} & \bf{-800} & \bf{-1600} & {TrajOpt-58} & {GPMP2-12} & {PRM} & {RRT-Connect} & RRT*\\ 
\midrule
success rate (\%) & 52.78 & 75 & 90.56 & 30.56 & 43.33 & 85 & 95.56 & 80.56 \\
average time (s) & 9.45 & 10.2 & 5.83 & 0.24 & 0.75 & 15.1 & 12.4 & 18.2 \\
standard deviation (s) & 3.45 & 2.35 & 4.213 & 0.12 & 0.23 & 4.03 & 4.35 & 2.37 \\
\bottomrule
\end{tabular}}
\end{threeparttable}
\end{table*}
%


The self comparison under the variation of the sample number $N$ among $\{400, 800, 1600\}$ in \Cref{tab:results:iiwa} shows a relatively large number (BN-MCO-1600) per step can elevate the success rate by 30\%-60\% with the broader information of the feasible space and improve the searching efficiency by 15\%-45\% with fewer learning steps under the information. It indicates that can BN-MCO can learn a better GMM that can attach the constraints more sophisticatedly from more samples per step though more samples means more computation resource for constraint check and model learning. 

\Cref{tab:results:iiwa} also shows BN-MCO-1600 gains the second highest success rate, compared to the other numerical and sampling planners. Though GPMP adopts the Gauss-Newton method to overcome some local minima faced by the TrajOpt using the trust-region method during trajectory optimization, some of its planned trajectories still get stuck in the obstacles with 33.3\% success rate when the planning scenario becomes narrow, as shown in \Cref{fig:p3}. So PRM resolves this issue and improves the success rate from 10\% to 30\% by introducing the construction, expansion and query phases so that the feasible paths can be found from the well-connected roadmap consists of the samples uniformly covering the feasible space, though it takes 14 more seconds to plan. Unlike the roadmap construction of PRM, RRT-Connect grows the rapidly random searching trees in bi-direction and pairs the bi-RRTs to gain the feasible paths with 20\% lower computation time compared to PRM. Though RRT* is asymptotically optimal version of RRT, tending to find the low cost results, its rewiring process consumes a significantly high computation resources so that its success rate is restricted by 20\% under the given maximum time. These results verify our incremental learning based on Monte-Carlo optimization can somehow improve the planning efficiency with acceptable success rate.

\section{CONCLUSIONS}

BN-MCO utilizes the Monte-Carlo sampling method to learn the distribution of a potential field online to plan the trajectory in a narrow workspace. 

(i) Our method learns the multi-modal via the sequential Monte-Carlo sampling of waypoints in the configuration space

(ii) BN-MCO adopt the online learning to incrementally minimize the KL-divergence between the true distribution of the potential field and the mixture distribution defined by the multi-modals

(iii) The bayes net, consisting of a bunch of related waypoints, can significantly reduce the amount of nodes and cut the redundant edges to improve the searching efficiency. 




\section*{ACKNOWLEDGMENT}

This work is supported by the Key R\&D Program of Zhejiang Province (2020C01025, 2020C01026), the National Natural Science Foundation of China (52175032), and Robotics Institute of Zhejiang University Grant (K11808).



\bibliographystyle{IEEEtran}
\bibliography{MCOMP_bib}

\end{document}